\documentclass{article}
\usepackage{amsmath,graphicx,mlspconf}
\usepackage{xcolor}
\usepackage{graphicx}
\usepackage{amsmath}
\usepackage{amssymb}
\usepackage{booktabs}
\usepackage{comment}
\usepackage{enumitem}
\usepackage{url}
\copyrightnotice{U.S.\ Government work not protected by U.S.\ copyright}

\copyrightnotice{979-8-3503-2411-2/25/\$31.00 {\copyright}2025 Crown}

\copyrightnotice{979-8-3503-2411-2/25/\$31.00 {\copyright}2025 European Union}

\copyrightnotice{979-8-3503-2411-2/25/\$31.00 {\copyright}2025 IEEE}

\toappear{2025 IEEE International Workshop on Machine Learning for Signal Processing, Aug.\ 31-- Sep.\ 3, 2025, Istanbul, Turkey}

\title{Subject Invariant Contrastive Learning for Human Activity Recognition}
\name{Yavuz Yarici  \enskip Kiran Kokilepersaud  \enskip Mohit Prabhushankar \enskip Ghassan AlRegib \enskip }

\address { OLIVES at the Center for Signal and Information Processing CSIP,\\ 
School of Electrical and Computer Engineering, Georgia Institute of Technology, Atlanta, GA, USA \\
\{yavuzyarici, kpk6, mohit.p, alregib\}@gatech.edu   }

\begin{document}
\ninept

%\begin{comment}

\twocolumn[{%

{ \large
\begin{itemize}[leftmargin=2.5cm, align=parleft, labelsep=2cm, itemsep=4ex,]
\vspace{1in}
\item[\textbf{Citation}]{Y. Yarici, K. Kokilepersaud, M. Prabhushankar, and G. Alregib, "Subject Invariant Contrastive Learning for Human Activity Recognition" in 2025 IEEE 35th International Workshop on Machine Learning for Signal Processing (MLSP), Istanbul, Turkey 2025.
}

\item[\textbf{Review}]{Date of Acceptance: June 24th 2025}

\item[\textbf{Codes}]{\url{https://github.com/olivesgatech/SICL}}

\item[\textbf{Bib}]  {@inproceedings\{yarici2025subject,\\
    title=\{Subject Invariant Contrastive Learning for Human Activity Recognition\},\\
    author=\{Yarici, Yavuz and Kokilepersaud, Kiran and Prabhushankar, Mohit and AlRegib, Ghassan\},\\
    booktitle=\{2025 IEEE 35th International Workshop on Machine Learning for Signal Processing (MLSP)\},\\
    address=\{Istanbul, Turkey\},\\
    publisher=\{IEEE\},\\
    year=\{2025\}\}}

\item[\textbf{Copyright}]{\textcopyright 2025 IEEE. Personal use of this material is permitted. Permission from IEEE must be obtained for all other uses, in any current or future media, including reprinting/republishing this material for advertising or promotional purposes, creating new collective works, for resale or redistribution to servers or lists, or reuse of any copyrighted component of this work in other works.}

\item[\textbf{Contact}]{

\{yavuzyarici, kpk6, mohit.p, alregib\}@gatech.edu \\\url{https://alregib.ece.gatech.edu/}\\}
\end{itemize}

}}]

%\end{comment}

\maketitle

\begin{abstract}
The high cost of annotating data makes self-supervised approaches, such as contrastive learning methods, appealing for Human Activity Recognition (HAR). Effective contrastive learning relies on selecting informative positive and negative samples. However, HAR sensor signals are subject to significant domain shifts caused by subject variability. These domain shifts hinder model generalization to unseen subjects by embedding subject-specific variations rather than activity-specific features. As a result, human activity recognition models trained with contrastive learning often struggle to generalize to new subjects. We introduce Subject‑Invariant Contrastive Learning (SICL), a simple yet effective loss function to improve generalization in human activity recognition. SICL re‑weights negative pairs drawn from the same subject to suppress subject‑specific cues and emphasize activity‑specific information. We evaluate our loss function on three public benchmarks: UTD-MHAD, MMAct, and DARai. We show that SICL improves performance by up to 11\% over traditional contrastive learning methods. Additionally, we demonstrate the adaptability of our loss function across various settings, including multiple self-supervised methods, multimodal scenarios, and supervised learning frameworks.
\end{abstract}

\section{Introduction} \label{sec:intro}\vspace{-1mm}

Human activity recognition (HAR) has broad applications in many areas, such as smart homes \cite{mehr2019human} and health monitoring \cite{siirtola2019personalizing}. However, these applications rely on massive data drawn from various sensors. Labeling this data presents a major challenge, as it is expensive and time consuming \cite{kokilepersaud2023focal}. Furthermore, labeling time series data, such as accelerometer signals, is even more demanding. It requires synchronized video recordings to accurately annotate the activities corresponding to each time point \cite{kaviani2025hierarchicalmultimodaldatadaily}.

Recently, self-supervised learning (SSL) has shown great effectiveness in learning representations without the need for expensive, manually annotated data. Specifically, contrastive self-supervised learning has proven to be effective across various domains, including computer vision \cite{chen2020simple, zbontar2021barlow, bardes2021vicreg}, natural language processing (NLP) \cite{gao2021simcse}, and sensor-based applications \cite{khaertdinov2021contrastive}.

The goal of contrastive learning is to learn a representation that maps semantically similar (positive) data points to nearby locations, while pushing semantically different (negative) data points farther apart. In computer vision, positive pairs are typically generated from an anchor data sample through image augmentations that preserve semantic content. Negative samples are simply drawn uniformly from the training data, without considering their informativeness for learning discriminative representations. This framework has been adapted for sensor-based human activity recognition by employing signal augmentations to form positive pairs during contrastive loss computation \cite{khaertdinov2021contrastive}. Negative samples are again drawn uniformly from the data. Recent methods have suggested more informative negative sampling strategies, such as hard negative sampling \cite{choi2023multimodal} and cross-model knowledge integration \cite{brinzea2022contrastive}. However, these methods do not consider distribution shifts in HAR data during the sampling of negatives. 

\begin{figure}
    \centering
    \includegraphics[width=0.95\linewidth]{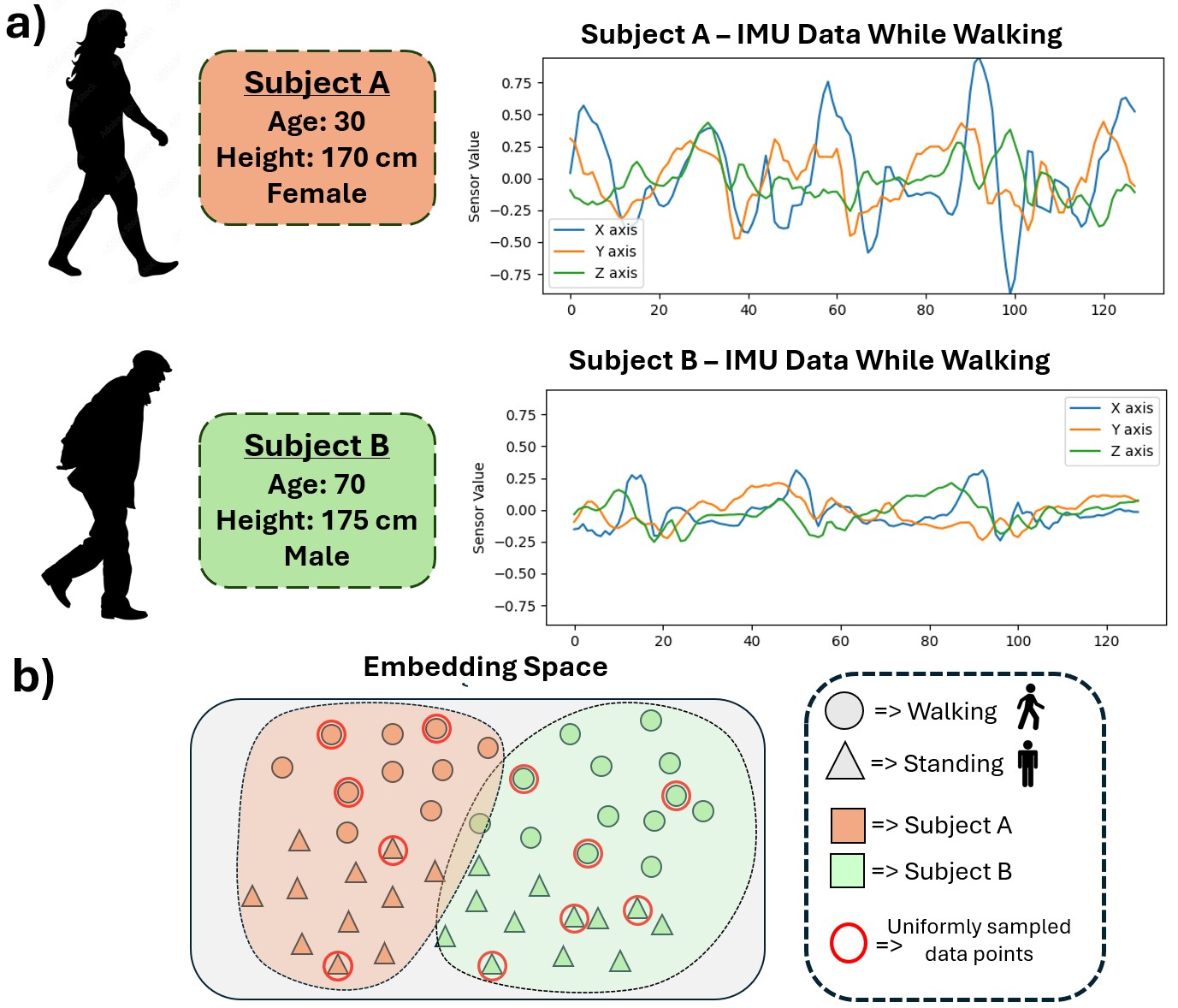}
    \caption{This figure demonstrates the variations in activity patterns captured from different subjects. a) Although both samples belong to the same walking activity, the data from adult Subject A shows considerably greater variation compared to the data from older Subject B due to inherent subject variability in human activity data. b) The data from two different subjects show distributional shifts in the embedding space. Traditional methods sample these points uniformly without considering distributional shifts. Our loss function considers these distributional shifts to learn subject-invariant features}
    \vspace*{-1.5em}
    \label{fig:walking_data}
\end{figure}

In HAR domain, the data collected from different subjects exhibit variations in their activity patterns. As shown in part a) of Figure 1, the data from adult Subject A shows considerably greater variation compared to the data from older Subject B. These subject-based variations introduce an intrinsic distribution shift. As illustrated in part b) of Figure 1, the data 
from two different activities shows distributional shifts based on subjects. Sampling negatives uniformly from this data hinders the model's ability to learn distinctive characteristics of the activities themselves, as models tend to capture subject-specific features.

The problem of distributional shifts is more significant when the model is applied to unseen target domains with poor generalization performance. If the testing data diverges from the training data due to subject heterogeneity, the performance of the model significantly decreases. For instance, a model that recognizes the activities of an adult may not generalize well to unseen data from an elderly person. Hence, addressing distributional shifts caused by subject heterogeneity is essential for developing robust human activity recognition systems that generalize effectively.

In this work, we propose a simple addition to the standard contrastive loss to address subject-based distributional shifts. We modify the standard contrastive loss function to effectively utilize subject labels. Our Subject-Invariant Contrastive Learning (SICL) loss function adjusts the weighting of same subject samples during contrastive learning, enforcing greater separation between samples from the same subject. This  encourages the model to learn distinctive characteristics of activities rather than subject-specific features, resulting in representations that are invariant to subject identity. We demonstrate that our loss function introduces subject invariance into the learned representations and can be integrated with traditional contrastive learning objectives. Furthermore, it is adaptable to various contrastive learning frameworks for HAR, including both supervised and multimodal settings.

In summary, we make the following contributions:

\begin{enumerate}[itemsep=0.5pt, topsep=1pt]

\item We introduce a simple and effective method to integrate subject labels during contrastive training, making the representations invariant to subject variations. 

\item We show that our approach outperforms traditional contrastive methods and, when combined with those methods, further boosts their performance by making the learned representations subject-invariant.

\item We show that our method can be used as a pre-training step for supervised training to increase generalization performance. 

\item We show that subject-invariant contrastive learning is also effective in supervised contrastive settings and recent multimodal HAR frameworks. 

\end{enumerate}

\vspace{-1mm}
\section{Related Works} \label{sec:related}\vspace{-1mm}

\subsection{Contrastive Learning and HAR}

Contrastive loss maps semantically similar (positive) samples closer in the embedding space while pushing dissimilar (negative) samples farther apart.  This approach has proven effective at extracting meaningful representations from HAR data. For instance, one prominent approach \cite{khaertdinov2021contrastive} adapts the SimCLR framework with a transformer-based encoder for HAR sensor data. It generates positive pairs through signal augmentation methods such as scaling, shuffling, jittering, and rotation. Similarly, \cite{choi2023multimodal} improves the learned representation by reweighting hard negatives that have different labels from the anchor but are projected nearby in embedding space. COCOA \cite{deldari2022cocoa} introduces a cross-modality contrastive framework that aligns sensor data from heterogeneous sources to improve robustness against missing data modalities. CMC-CMKM \cite{brinzea2022contrastive} combines contrastive learning with cross-modal knowledge mining to enhance information transfer between modalities.

\subsection{Distribution Shift Problem in Human Activity Data}

A number of techniques have been proposed to address the distribution shift problem in Human Activity Recognition (HAR) by learning invariant representations that can generalize across different subjects. For instance, one prominent study in \cite{jimale2023subject} investigates how subject variability affects sensor-based activity recognition. They show that models trained on one demographic experience significant performance drops when applied to others. Similarly, \cite{xiong2024generalizable} introduces a categorical concept invariant learning framework to enforce invariance at both the feature and output levels. This leads to improved cross-domain generalization. In another approach, \cite{lu2024diversify} proposes a min–max adversarial strategy to automatically discover latent sub-domain shifts to improve out-of-distribution detection performance. Furthermore, \cite{qian2021latent} presents a latent independent excitation mechanism that disentangles user-specific factors from activity-related features. This gives robust cross-person generalization performance without requiring target domain data. These works all identify ways to use variants of machine learning methods to enable models to perform well in the target domain. However, they fundamentally differ from our approach, as they do not offer a framework that operates within a self-supervised contrastive learning paradigm.

 %Furthermore, \cite{hao2021invariant} leverages a combination of meta-learning and multi-task learning to extract invariant features from sensor data collected across diverse domains, which boosts cross-domain performance.
\vspace{-1mm}
\section{Subject Variability in HAR} \label{sec:subject_variability}\vspace{-1mm}

We first conduct an analysis to investigate whether the data acquired during human activities exhibits a domain shift problem due to subject-based variability. For this analysis, we use the DARai \cite{kaviani2025hierarchicalmultimodaldatadaily} dataset, as it consists of daily activities collected in a natural, real-world setting from different subjects. This feature of dataset allow us to see the natural variations in how different subjects perform the same activities. We compute two key signal features: spectral entropy and dominant frequency. These features are calculated on a per-channel basis and then averaged over all channels. The resulting values are then grouped by subject. We conduct one-way ANOVA across subjects to determine if the differences across subjects are significant. A one-way ANOVA across subjects revealed statistically significant differences:
\begin{align*}
    F_{\text{Spectral Entropy}} &= 3.031,\quad p = 4.047\times10^{-5}, \\
    F_{\text{Dominant Frequency}} &= 2.333,\quad p = 1.768\times10^{-3}.
\end{align*}
The $p$-values from the ANOVA tests indicate that spectral entropy and dominant frequency differ significantly across subjects.

%In addition, we perform an activity-based analysis. We group and compare samples of each activity by subject. This analysis reveals substantial statistical differences (with p-values as low as $10^{-9}$). This demonstrates that even within the same activity, subjects exhibit distinct signal characteristics.

% To obtain an overall measure of subject-specific variability across all activities, we combine the individual activity-level p-values using Fisher's method. This yields the following combined statistics:
% \begin{align*}
%     \chi^2(54) &= 763.247,\quad p \approx 0, \quad \text{(Spectral Entropy)}, \\
%     \chi^2(54) &= 530.484,\quad p \approx 0, \quad \text{(Dominant Frequency)}.
% \end{align*}
% Combined results provide strong evidence that the signal characteristics of inertial sensor signals vary significantly across subjects. The observed subject-specific differences, reflected in both spectral entropy and dominant frequency.

% \begin{figure}[b]
%     \centering
%     \includegraphics[width=0.9\linewidth]{Figures/seen_unseen_comparison.png}
%     \caption{This plot compares top-1 accuracies of seen subject and unseen subject setting for different data modalities.}
%     \label{fig:seensubject_comparision}
% \end{figure}

\begin{figure}[tbh]
    \centering
    \includegraphics[width=0.95\linewidth]{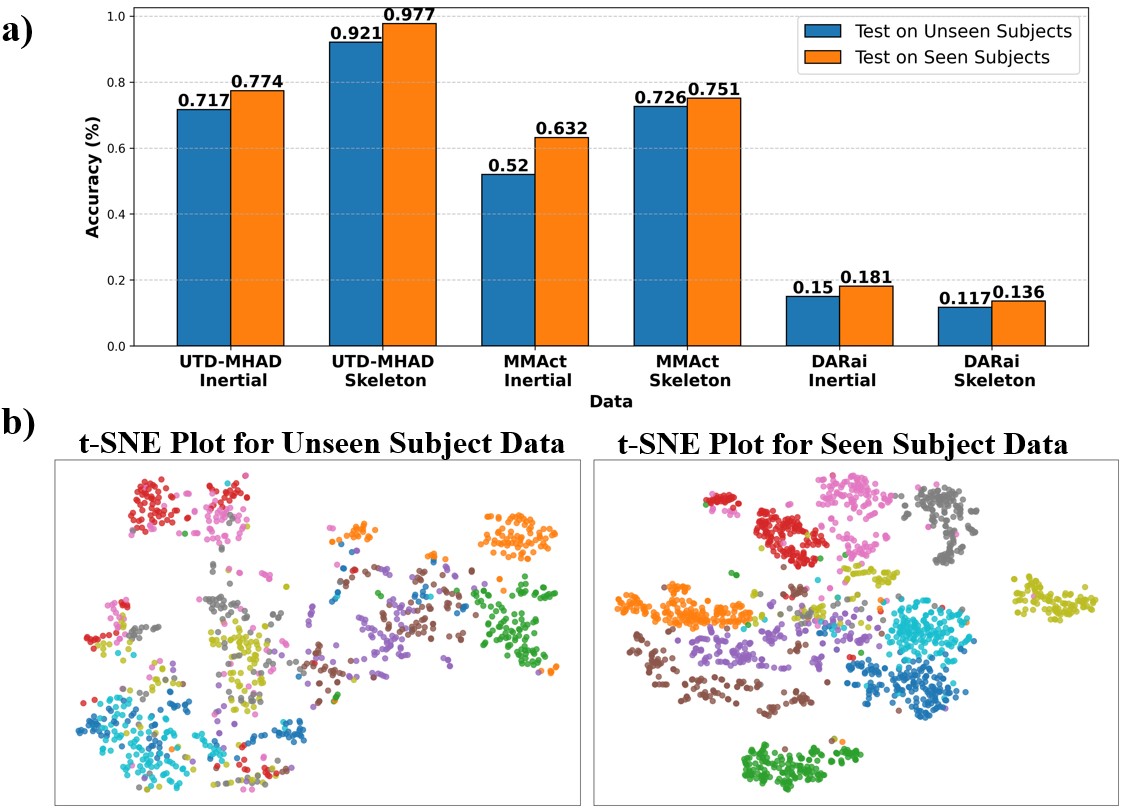}
    \caption{a) This plot compares top-1 accuracies of seen subject and unseen subject setting for different data modalities with SimCLR training. b) This figure shows the t-SNE plot of seen and unseen subject settings for inertial data. In the seen subject setting.
    }
    \label{fig:seensubject_comparision}
\end{figure}

Additionally, we conduct an analysis of the effect of domain shift for contrastive learning models for HAR data. We compare two experimental settings. In the seen-subject setting, the training and testing data are randomly shuffled, allowing the same subjects to appear in both sets. In contrast, in the unseen-subject (cross-subject) setting, training and testing data are separated based on subjects. This corresponds to the traditional evaluation methods for HAR. It ensures that the test set comprises entirely new subjects not present during training. The results of SimCLR\cite{chen2020simple} for different datasets and data modalities are shown in part a) of Figure \ref{fig:seensubject_comparision}. The performance for activity recognition in the unseen-subject setting is lower than in the seen-subject setting. This suggests that standard contrastive loss is not able to learn distinctive activity features that can generalize well to unseen subjects. We also show the t-SNE plot of seen and unseen subject settings in part b) of Figure \ref{fig:seensubject_comparision}. In the seen subject setting, the test data are clustered according to activity labels. This shows that the representation is able to distinguish differences between activities,

% \begin{figure}[t]
%   \centering
%   \includegraphics[width=0.9\linewidth]{Figures/tsne.jpg}
%   \caption{ This figure shows the t-SNE plot of seen and unseen subject settings. In the seen subject setting, the test data is significantly more clustered according to activity labels.}
%   \label{fig:tsne plot}
% \end{figure}

Additionally, we look at the distribution of cosine similarity pairs for inertial data from DARai\cite{kaviani2025hierarchicalmultimodaldatadaily} dataset. In Figure \ref{fig:cosinesimilaritycompare}, we show the cosine similarity distribution of pairs computed across all samples as well as for samples within the same subject, denoted as intra-subject pairs. The results show that, for standard contrastive training, most samples exhibit higher cosine similarities for intra-subject pairs compared to all pairs, confirming that there is a distributional shift in the learned representation space with respect to subject distribution.

\vspace{-1mm}
\section{Methodology} \label{sec:method}\vspace{-1mm}

\begin{table*}[htb]
\footnotesize
\centering
\begin{tabular}{@{}ccccccc@{}}
\toprule

                              & UTD-MHAD & UTD-MHAD & MMAct    & MMAct    & DARai    & DARai         \\
                              & Inertial & Skeleton & Inertial & Skeleton & Inertial & Foot Pressure \\ \toprule
SimCLR                        & 0.7173   & 0.9211   & 0.5202   & 0.7257   & 0.1498   & 0.1172        \\
HCL                           & 0.7210   & 0.9240   & 0.5208   & 0.7310   & 0.1510   & 0.1247        \\
SICL                & 0.7359   & 0.9304   & 0.5326   & 0.7413   & \textbf{0.1615  } & 0.1316        \\
SimCLR+SICL        & \textbf{0.7391}   & \textbf{0.9416}   & \textbf{0.5333 }  &\textbf{ 0.7436}   & 0.1610   & \textbf{0.1342}        \\

 \midrule
Barlow Twins                  & 0.7260   & 0.9026   & 0.5254   & 0.7291   & 0.1443   & 0.1218        \\
Barlow Twins +SICL  & \textbf{0.7419}   & \textbf{0.9327}   & \textbf{0.5364}   & \textbf{0.7338}   & \textbf{0.1583}   &\textbf{ 0.1325 }       \\
\midrule
VICReg                        & 0.7238   & 0.9392   & 0.5361   & 0.7335   & 0.1516   & 0.1181        \\
VICReg +SICL       & \textbf{0.7446 }  & \textbf{0.9453}   & \textbf{0.5426}   & \textbf{0.7497}   & \textbf{0.1615 }  & \textbf{0.1289}        \\

           \bottomrule
\end{tabular}
\caption{ This table shows the Top-1 accuracy of Subjcet Invariant Contrastive Learning (SICL) within a linear evaluation setting. We bold the highest performance in each section.} 
\label{tab:linear_eval_uni}
\end{table*}

We modify the traditional contrastive loss function to effectively utilize subject labels. Our loss function selectively samples contrastive negatives based on their subject labels. This encourages the model to learn distinctive characteristics of activities rather than subject-specific features. 

Similar to \cite{kokilepersaud2024taxes, kokilepersaud2024hex}, we decompose our negatives into two sets. One set consists of negatives from the same subject, and the other consisting of negatives from different subjects. We then regularize the same-subject negatives to encourage the learning of subject-invariant representations. Our proposed approach is discussed in detail below:

\subsection{Subject Invariant Contrastive Loss for Unimodal Data}

In sensor-based human activity recognition, signals collected from wearable sensors are typically represented as multivariate time series data. At each timestamp \(t\), the input signal is defined as

{\small
\[
x_t = [x_t^1,\, x_t^2,\, \dots,\, x_t^S] \in \mathbb{R}^S, \quad
X = [x_1,\, x_2,\, \dots,\, x_T],
\]
}

\noindent where \(S\) is the number of channels and \(T\) is the number of timesteps aggregated into a time window. The goal of human activity recognition is to associate each time-window \(X\) with the correct output label \(y \in Y\).

Contrastive learning for time series data employs a framework in which each unlabeled time series \(X\) is subject to random augmentations to generate multiple views. Two sets of random augmentations, denoted by \(t\) and \(t'\), are applied to each input signal \(X\) in a batch. This process yields the augmented views \(t(X)\) and \(t'(X)\), which form a positive pair. These views are then processed by an encoder network \(f(\cdot)\), which transforms them into latent representations \(h_i = f(t(X))\) and \(h_j = f(t'(X))\). Then, a projection network \(g(\cdot)\) maps these latent representations to embeddings \(z_i = g(h_i)\) and \(z_j = g(h_j)\).

\begin{figure}
    \centering
    \includegraphics[width=0.7\linewidth]{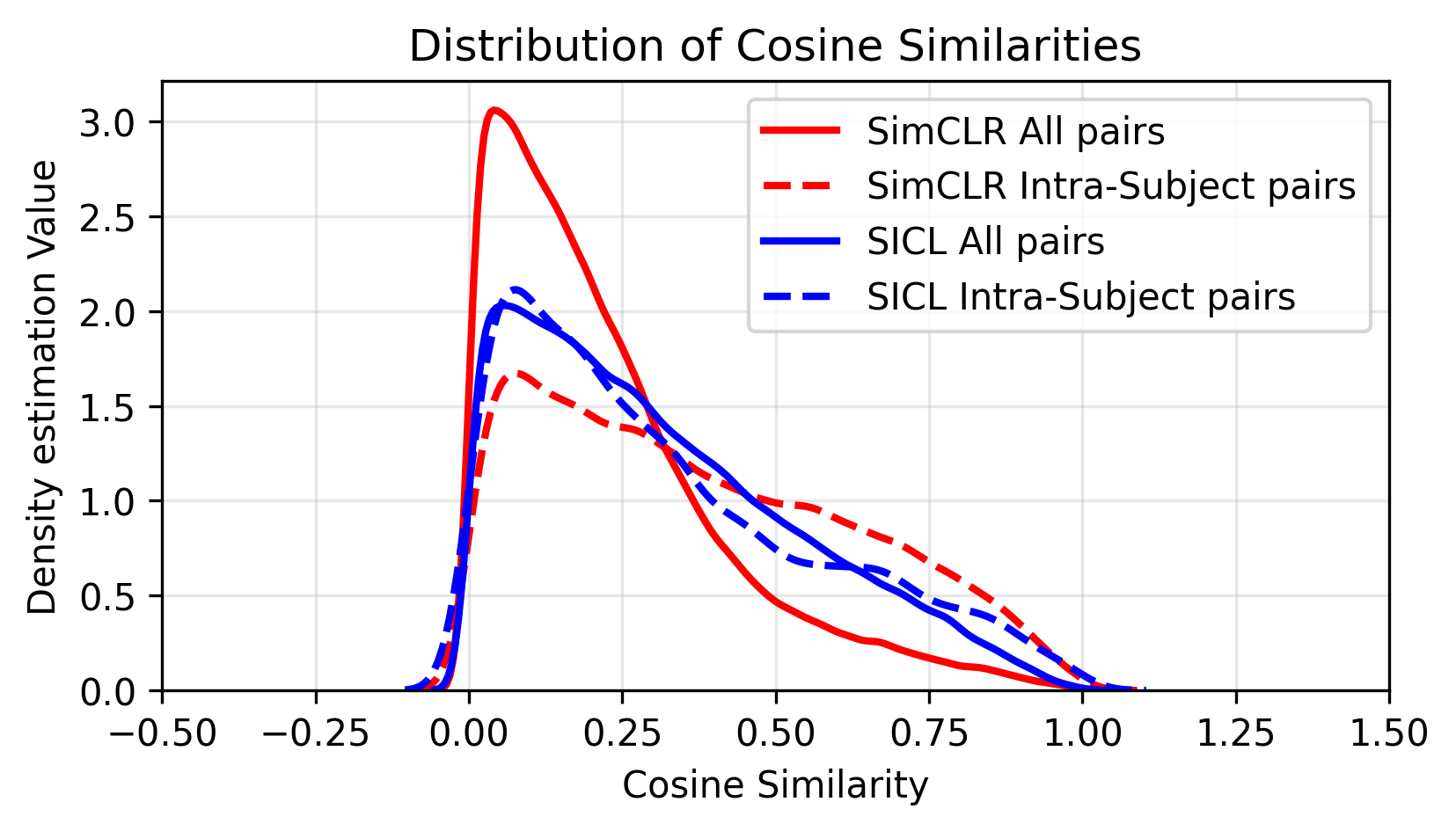}
  \caption{This figure shows the distribution of cosine similarity scores.} 
  \label{fig:cosinesimilaritycompare}
\end{figure}

By treating \(z_i\) and \(z_j\) as a positive pair, the objective is to pull their embeddings closer together in the embedding space. The remaining samples in the batch form the negative set, denoted as \(A(i)\). These samples belong to different instances and the objective is to push their embeddings farther apart. This objective is formalized through the following loss function:
{\small
\begin{equation}
    L_{\text{NCE}} = -\sum \log \left( \frac{\exp(z_i \cdot z_j / \tau)}{\sum_{a \in A(i)} \exp(z_i \cdot z_a / \tau)} \right),
\end{equation}
}
where \(\tau\) is a temperature parameter that scales the dot products.

While the standard contrastive loss encourages learning representations that are invariant to augmentations and discriminative across instances, it does not account for the distributional shifts due to subject variations. To address this limitation, we refine the loss function by decomposing the negative set \(A(i)\) into two subsets: negatives from the same subject, denoted as \(S(i)\), and negatives from different subjects. Furthermore, we introduce a weighting function \(Q_{Si}\) that assigns different weights to negatives from the same subject. \(Q_{Si}\) is calculated batch-wise, similar to \cite{kokilepersaud2024taxes}, and is constructed by normalizing the exponentiated cosine similarities of same-subject negatives, summing these values, and then dividing by their average. Our final subject invariant loss is formulated as:

{\small
\begin{equation}
    L_{\text{SICL}} = -\sum_i \log \left( \frac{\exp\left(\frac{z_i \cdot z_j}{\tau}\right)}{D_i} \right),
\end{equation}
}

where \(D_i\) is defined as

{\small
\begin{equation}
    D_i = Q_{Si} \sum_{s \in S(i)} \exp\left( \frac{z_i \cdot z_s}{\tau} \right) + \sum_{k \notin S(i)} \exp\left( \frac{z_i \cdot z_k}{\tau} \right).
\end{equation}
}

The weighting function reweights negatives from the same subject. It increases the spread among these negatives while reducing intra-subject similarities. As shown in Figure \ref{fig:cosinesimilaritycompare}, the distributions of cosine similarities for all pairs and intra-subject pairs are much closer to each other for SICL compared to SimCLR. This adjustment encourages the model to learn semantic differences between activities rather than subject based differences. This makes the learned representation subject invariant. 

\begin{table}[htb]
\small
\centering
\resizebox{\columnwidth}{!}{%
\begin{tabular}{@{}cccc@{}}
\toprule
     & \shortstack{UTD-MHAD\\(Inertial + Skeleton)} 
       & \shortstack{MMAct\\(Inertial + Skeleton)}      
       & \shortstack{DARai\\(Inertial + Foot Pressure)} \\ 
\toprule
CMC                          & 0.951 & 0.810 & 0.239 \\
CMC SICL          & \textbf{0.959} & \textbf{0.8289} &\textbf{ 0.254} \\

\midrule
CMC-CMKM                    & 0.953    &  0.826     &   0.241    \\
CMC-CMKM SICL    & \textbf{0.957}    & \textbf{0.831 }    &   \textbf{0.243 }   \\

\bottomrule
\end{tabular}
}
\caption{This table shows the Top-1 accuracy of Subject Invariant Contrastive Learning (SICL) for multimodal settings within a linear evaluation setting.}
\label{tab:linear_eval_multi}
\end{table}

\subsection{Subject Invariant Supervised Contrastive Loss}

By leveraging label information, the supervised contrastive loss \cite{khosla2020supervised} treats every pair of samples with the same label as a positive pair. For each sample \(z_i\), the positive set \(P(i)\) is defined as all other samples that share the same label, while the negative set \(A(i)\) consists of samples from different classes. The objective is to pull together the embeddings of samples within the same class and push apart the embeddings of samples from different classes. This objective is formalized by the following loss function:

\begin{equation}
    L_{supcon} = -\sum_{i=1}^{N} \frac{1}{|P(i)|} \sum_{p \in P(i)} \log \frac{\exp\left(\frac{z_i \cdot z_p}{\tau}\right)}{\sum_{a \in A(i)} \exp\left(\frac{z_i \cdot z_a}{\tau}\right)}
\end{equation}

Here, \(P(i)\) denotes the set of indices corresponding to samples that share the same label as \(z_i\), and \(A(i)\) denotes the set of indices corresponding to samples from different classes. The temperature parameter \(\tau\) scales the dot products.

Similar to the unsupervised case, we further refine the supervised contrastive loss by decomposing the negative set \(A(i)\) into two subsets and introducing a weighting function \(Q_{Sij}\) to reweight negatives from the same subject. Our final subject-invariant supervised contrastive loss is formulated as:

\begin{equation}
    L_{SISup} = -\sum_{i=1}^{N} \frac{1}{|P(i)|} \sum_{p \in P(i)} \log \frac{\exp\left(\frac{z_i \cdot z_p}{\tau}\right)}{D_i}
\end{equation}

\begin{table*}[htb]
\scriptsize
\centering
\begin{tabular}{@{}ccccccc@{}}
\toprule

                              & UTD-MHAD & UTD-MHAD & MMAct    & MMAct    & DARai    & DARai         \\
                    Backbone Pre-training          & Inertial & Skeleton & Inertial & Skeleton & Inertial & Foot Pressure \\ \toprule
No Pre-training  & 0.7592 & 0.9466 & 0.6117 & 0.8018 & 0.1683 & 0.1401 \\
SimCLR          & 0.7564 & 0.9472 & 0.6213 & 0.7944 & 0.1675 & 0.1401 \\
HCL             & 0.7613 & 0.9447 & 0.6288 & 0.7964 & 0.1702 & 0.1438 \\
SICL  & \textbf{0.7772} & \textbf{0.9535} & \textbf{0.6557} &\textbf{ 0.8212} & \textbf{0.1793} & \textbf{0.1513} \\

           \bottomrule
\end{tabular}
\caption{ This table shows the Top-1 accuracy for supervised classification results for self-supervised pretraining with Subject Invariant Contrastive Learning (SICL) on UTD-MHAD, MMact and DARai dataset. We bold the highest performance in each dataset.}
\label{tab:pretraining}
\end{table*}

\begin{table*}[htb]
\scriptsize
\centering
\begin{tabular}{@{}ccccccc@{}}
\toprule

                              & UTD-MHAD & UTD-MHAD & MMAct    & MMAct    & DARai    & DARai         \\
                    Backbone Pre-training          & Inertial & Skeleton & Inertial & Skeleton & Inertial & Foot Pressure \\ \toprule
Cross-Entropy Loss   & 0.7592   & 0.9466   & 0.6117   & 0.8018   & 0.1683   & 0.1401        \\
SupCon               & 0.7648   & 0.9526   &  0.6215        & 0.8141   & 0.1715   & 0.1510        \\
SupHCL               & 0.7666   & 0.9570   &   0.6188       & 0.8187   & 0.1742   & 0.1502        \\
SI-SupCon  & \textbf{0.7842}   & \textbf{0.9637}   &   \textbf{0.6427}       & \textbf{0.8205}   & \textbf{0.2031 }  &\textbf{ 0.1592}        \\

           \bottomrule
\end{tabular}
\caption{ This table shows the Top-1 accuracy for supervised classification results for Subject Invariant Supervised Contrastive Learning on UTD-MHAD, MMact and DARai dataset. We bold the highest performance in each dataset.}
\label{tab:supervised_ssl}
\end{table*}

\subsection{Subject Invariant Contrastive Loss for Multi-modal Data}

To implement our subject-invariance loss on multimodal data, we adopt two contrastive multiview methods: Contrastive Multiview Coding (CMC) \cite{tian2020contrastive} and Contrastive Multiview Coding with Cross-Modal Knowledge Mining (CMC-CMKM) \cite{Brinzea_2022}.

In the CMC framework, both positive and negative pairs consist of samples from two distinct data modalities derived from the same activity session. The loss formulation is analogous to that presented in \cite{chen2020simple}, with the key difference being that the pairs are drawn from different modalities. The objective function for CMC is formalized as follows:

{\scriptsize
\begin{equation}
L_{CMC} = -\sum \log\left(\frac{e^{z_i^k \cdot z_i^m/\tau}}{\sum_{a\in A(ij)} e^{z_i^k \cdot z_a^m/\tau}}\right)
\end{equation}
}

where \(z_i^k\) denotes the representation from data modality \(k\) and \(z_i^m\) and \(z_a^m\) denote the representations from data modality \(m\).

Similar to the unimodal case, we apply our subject-based negative decomposition strategy. Our subject invariant CMC loss function is formulated as follows:

{\scriptsize
\begin{equation}
L_{SI\text{-}CMC} = -\sum \log\left(\frac{e^{z_i^k \cdot z_i^m/\tau}}{Q_{Sij} \sum_{s\in S(i)} e^{z_i^k \cdot z_s^m/\tau} + \sum_{k\notin S(i)} e^{z_i^k \cdot z_k^m/\tau}}\right)
\label{CI-CMC}
\end{equation}
}

CMC-CMKM extends this approach by first training separate encoders to guide cross-modal learning through mining of positive and negative pairs from unimodally trained encoders. Subsequently, the CMC loss is applied using these mined positives and negatives. We use the same decomposition as in Equation \ref{CI-CMC} to enforce subject invariance in CMC-CMKM.

\vspace{-1mm}
\section{Experiments and Results} \label{sec:results}\vspace{-1mm}
\subsection{Datasets}
We evaluate the performance of our proposed approaches on three benchmark multimodal datasets: UTD-MHAD \cite{chen2015utd}, MMAct \cite{Kong_2019_ICCV}, and DARai \cite{kaviani2025hierarchicalmultimodaldatadaily}.

\textbf{UTD-MHAD \cite{chen2015utd}} consists of recordings from 10 subjects performing 27 distinct activities, with each activity repeated over four trials. We use the skeleton and inertial data modalities from UTD-MHAD. 

\textbf{MMAct \cite{Kong_2019_ICCV}} consists of recordings from 20 subjects performing 36 distinct activities across various scenes. We use the skeleton and inertial data modalities from MMAct.

\textbf{DARai \cite{kaviani2025hierarchicalmultimodaldatadaily}} consists of recordings from 50 subjects performing various activities in kitchen, dining room, living room, and office environments. We use a subset of the dataset consisting of 20 subjects. We chose the DARai dataset because its rich, multimodal recordings of natural daily activities across diverse real-world settings provide a more challenging benchmark than other existing HAR datasets. DARai has 20 different data modalities. In this work, we use IMU and foot pressure data modalities.

%Three-dimensional joint coordinates were captured using a Kinect camera, providing precise spatial measurements, while a wearable device equipped with an accelerometer and gyroscope simultaneously recorded inertial data. To maintain consistency with the original evaluation protocol, we used data from odd-numbered subjects for training and data from even-numbered subjects for testing, reporting the test accuracy based on this split.

%For the skeleton data, we utilize the 2D keypoints provided in the challenge version of the dataset. We adhere to a cross-subject evaluation protocol, employing samples from the first 16 subjects for training while using the remaining subjects' data for testing.

%The dataset employs a three-level hierarchical annotation scheme to capture activities at varying levels of granularity. In our experiments, we utilize the Level 1 and Level 2 labels, corresponding to 18 high-level categories and 44 unique fine-grained categories, respectively. We focus on a subset of the dataset consisting of 20 subjects collected during the Phase 1 stage. Two modalities are used: inertial data and foot pressure data. Inertial data are collected from IMU sensors attached to the subjects' arms, while foot pressure data are recorded using insole pressure sensors placed inside the subjects' shoes. To maintain consistency with the original evaluation protocol, we follow a cross-subject evaluation strategy using the same training and testing splits as provided in the original dataset.

\subsection{Implementation Details}

For the inertial and insole data encoders, we use three layers of 1D-CNNs followed by transformer self-attention layers, as proposed in \cite{khaertdinov2021contrastive}. For the skeleton encoder, we use the lightweight convolutional co-occurrence feature learning network introduced in \cite{li2018co}. 

We resample all input sequences to 100 time steps across all data modalities. Additionally, we normalize the joint positions in each skeletal sequence relative to the first frame, following standard normalization protocols. We use the Adam optimizer with a learning rate of 0.001. We pretrain backbones for 300 epochs and linear layers for 100 epochs. We use the same temperature parameters as in \cite{brinzea2022contrastive}. We report the top-1 average accuracy computed class-wise.

\subsection{Evaluation}\vspace{2mm}
\subsubsection{Linear Evaluation}\vspace{-1mm}
\label{sec:linear evaluation}
To evaluate the quality of the representations learned by our subject-invariant contrastive method, we employ a linear evaluation protocol. In this procedure, the weights of the pre-trained encoder are frozen, and a linear classification layer is attached on top. This setup ensures that the downstream performance is solely attributable to the learned representations, without any fine-tuning of the encoder parameters.

We compare our approach against two baseline methods: SimCLR \cite{chen2020simple} and the hard negative contrastive loss \cite{robinson2020contrastive}. The evaluation is conducted in a unimodal setting, and the detailed results are presented in Table \ref{tab:linear_eval_uni}. Our findings indicate that the proposed method outperforms both SimCLR and the hard negative contrastive loss. Moreover, when our subject-invariant loss is applied on top of existing loss functions such as \cite{chen2020simple}, Barlow Twins \cite{zbontar2021barlow}, and VICReg \cite{bardes2021vicreg}, we observe consistent improvements in performance. These results demonstrate the effectiveness of incorporating subject invariance into contrastive learning.

Furthermore, we extend our proposed loss function to multimodal contrastive learning by incorporating it into the CMC \cite{tian2020contrastive} and CMC-CMKM \cite{Brinzea_2022} frameworks. The multimodal linear evaluation results, presented in Table \ref{tab:linear_eval_multi}, demonstrate that our subject-invariant contrastive loss function outperforms the baseline methods in multimodal settings. These findings underscore the effectiveness of our approach in improving generalization performance of our method across both unimodal and multimodal settings.

\subsubsection{Pretraining with Subject Invariant Loss}
To demonstrate the effectiveness of subject-invariant learning, we first pretrain our encoders using the proposed method. After pretraining, a linear classification layer is attached, and the entire model is subsequently fine-tuned using a cross-entropy classification loss. We compare the performance of models initialized with our subject-invariant pretraining against those initialized with no pretraining, SimCLR pretraining, and Hard Negative Contrastive Learning (HCL).
The experimental results, shown in Table \ref{tab:pretraining}, indicate that models pretrained with our subject-invariant loss consistently achieve higher classification accuracy. This improvement suggests that incorporating subject invariance during pretraining enables the encoder to learn more robust and class-discriminative representations, thereby facilitating better performance on downstream tasks.

\subsubsection{Subject Invariant Supervised Contrastive Loss}

As described in Section \ref{sec:method}, we apply our loss function to the supervised contrastive learning setting as well. To evaluate the quality of the representations learned under this framework, we employ a linear evaluation protocol analogous to the unsupervised case. The encoder is pretrained using our subject-invariant supervised loss, after which its weights are frozen and a linear classification layer is trained using a cross-entropy loss. We compare our approach with models trained in a fully supervised manner using cross-entropy loss, standard supervised contrastive loss, and supervised hard negative contrastive loss. Table \ref{tab:supervised_ssl} summarizes the results, demonstrating that our method consistently outperforms these baseline approaches. These findings indicate that incorporating subject invariance into supervised contrastive learning enhances the quality of the learned representations, even in the presence of label supervision.

\begin{comment}
    
\subsubsection{Intra Subject Variations for Subject Invariant Contrastive Loss}
As an ablation study we also tried using available labels about the distributional shift in the data that affects intra-subject variability.  Intra-subject variability occurs when a given activity performed by the same subject differently due to conditions of the performed activity. We use scene labels in mmact and DARai dataset and task labels from DARai with our loss function to decompose negatives. Similar to section \ref{sec:linear evaluation}, we use linear evaluation method to test our method. The evaluation is conducted in a unimodal setting, and the detailed results are presented in Table \ref{tab:linear_eval_uni_intrasubject}. Our findings indicate that subject invariant loss works using other factors that causes intra-subject differences. 
\input{Tables/linear_eval_intrasubject}

\end{comment}

 \vspace{-1mm}
\section{Conclusion}\vspace{-1mm} \label{sec:conclusion}
In this work, we introduce a contrastive loss function to leverage subject labels to learn subject-invariant representations. Through extensive evaluation, we show our Subject-Invariant Contrastive Learning (SICL) approach consistently improves generalization to unseen subjects in both unimodal and multimodal settings. Moreover, we show that our approach serves as a pretraining strategy for downstream supervised tasks and integrates seamlessly with the supervised contrastive loss. These results highlight the importance of incorporating subject invariance into self-supervised learning frameworks to enhance the robustness and applicability of HAR systems in real-world scenarios.

\subsection*{Acknowledgment}
This work is partially supported by a gift from Amazon Lab 126. Special thanks to Dr. Mashhour Solh. 

% References should be produced using the bibtex program from suitable
% BiBTeX files (here: strings, refs, manuals). The IEEEbib.bst bibliography
% style file from IEEE produces unsorted bibliography list.
% -------------------------------------------------------------------------
\bibliographystyle{IEEEbib}
\bibliography{strings,refs}

\end{document}